\documentclass[runningheads]{llncs}
\usepackage[T1]{fontenc}
\usepackage{graphicx}
\usepackage{bm}
\usepackage{tabularx}
\usepackage{amsmath, amsfonts}
\DeclareMathOperator*{\argmax}{arg\,max}
\usepackage{xcolor}

\begin{document}
\title{Bayesian Evidence Accumulation for Graph-Based Learning in
Dynamic Data-Driven Applications Systems
}
\titlerunning{B-EAGLe in DDDAS}
%
\author{Moises Ponce\inst{1} \and Sarah Harkins Dayton\inst{2} \and
Ioannis Schizas\inst{3} \and
Vasileios Maroulas \inst{4}}

\authorrunning{M. Ponce et al.}
%
\institute{University of Tennessee, Knoxville TN 37996, USA
\email{mponce@vols.utk.edu}\\
\and
University of Tennessee, Knoxville TN 37996, USA
\email{sharkin5@vols.utk.edu}\\
\and
US Army Combat Capabilities Development Command, Army Research Laboratory, Adelphi MD 20783, USA
\email{ioannis.d.schizas.civ@army.mil}\\
\and
University of Tennessee, Knoxville TN 37996, USA
\email{vasileios.maroulas@utk.edu}}
\maketitle              
\begin{abstract}


Graph Neural Networks (GNNs) are widely used in modern machine learning, yet most existing frameworks remain fundamentally static, limiting their applicability in Dynamic Data Driven Application Systems (DDDAS) where observations arrive sequentially under uncertainty. We introduce B-EAGLe (Bayesian Evidence Accumulation for Graph-Based Learning), a probabilistic framework inspired by CBGT-Net (cortico-basal ganglia-thalamic network) that reformulates graph classification as a sequential evidence accumulation process. B-EAGLe combines a GNN encoder with a Bayesian accumulator that incrementally updates posterior beliefs over class labels as new observations arrive, committing to a decision only when sufficient confidence is achieved. Experiments on CIFAR10 demonstrate competitive performance together with principled confidence estimation under sequential observations.

\end{abstract}
\section{Introduction}
    Graph-structured data arises naturally across diverse domains, including social networks, biological and chemical systems, and computer vision \cite{fan2019graphsocial,li2023surveysocial,pavlopoulos2011using,lesne2006complexbio}, often exhibiting dynamic or spatiotemporal structure \cite{sahili2023spatio,zheng2025survey}. Graph Neural Networks (GNNs) have become the de facto method for modeling such complex data in classification tasks (e.g., graph-level and node-level classification) \cite{khoshraftar2024survey,xia2021graph}.
    GNNs have evolved considerably to address dynamic settings
    ; however, the underlying decision paradigm has remained unchanged: given an input, produce a single output. This static framing assumes that all relevant information is available at inference time, an assumption that breaks down when observations arrive sequentially and decisions must be made under uncertainty.

    Despite their considerable progress, the application of GNNs within the Dynamic Data Driven Applications Systems (DDDAS) paradigm \cite{darema2005dddas} remains an open direction. DDDAS entails the ability of a computational model to dynamically incorporate streaming data from an instrumentation system at execution time and, in turn, to govern its own measurement and data acquisition process. This paradigm has enabled real-time decision-making in applications such as unmanned vehicle control, sensor selection, and natural disaster forecasting \cite{ouyang2007dynamic,madey2007enhanced,schizas2015dynamic,cencerrado2011prediction}.

    Graph-structured representations arise naturally in DDDAS environments, where entities such as sensors, controllers, and agents form relational networks that evolve over time. GNNs have been applied to forecasting and skeleton-based action recognition \cite{Jiang_2022,zhang2018graphedgeconvolutionalneural}. GNNs are therefore well-positioned to serve as the computational backbone of a DDDAS application, provided the decision-making paradigm is extended to support sequential, streaming observations.

    To address the aforementioned challenges, this paper presents the B-EAGLe (Bayesian Evidence Accumulation for Graph-Based Learning), a framework that reframes graph learning-based classification tasks as a sequential, confidence-driven decision-making process under the DDDAS paradigm. B-EAGLe sequentially receives observations and accumulates evidence. Decisions are deferred until a sufficient confidence threshold is achieved. This is done using a probabilistic Bayesian belief update that maintains a posterior distribution over a set of class labels across the streamed observations from connected measurement instruments.

    The rest of the paper is organized in the following manner: In Section~\ref{sec:prelim}, a discussion on graph preliminaries is provided and how graph classification can operate under the DDDAS paradigm. The problem formulation and the proposed architecture are described in Section~\ref{sec:prob}. In Section~\ref{sec:case}, B-EAGLe is executed on four datasets as sequentially streamed observations and the subsequent results are shown. Finally, this paper concludes with Section~\ref{sec:conclusion}, where conclusions are provided and a brief discussion of future work is given.

\section{Preliminaries and Related Works}
    \label{sec:prelim}
    \subsection{Graph Theory Notation}
        Define a graph as $\mathcal{G} = (\mathcal{V}, \mathcal{E}, \bm{X}, \bm{E})$, where $\mathcal{V}$ is the set of nodes, $\mathcal{E}$ is the set of edges such that $\mathcal{E} \subseteq \mathcal{V} \times \mathcal{V}$. The node feature matrix, $\bm{X} \in \mathbb{R}^{|\mathcal{V}| \times d_\mathcal{V}}$, contains each node's feature vector, where $|\mathcal{V}|$ is the cardinality of $\mathcal{V}$ and $d_\mathcal{V}$ is the dimension of the node features. The edge feature matrix, $\bm{E} \in \mathbb{R}^{|\mathcal{E}| \times d_\mathcal{E}}$ contains each edge's feature vector, where $|\mathcal{E}|$ is the number of edges and $d_\mathcal{E}$ is the dimension of the edge features. The adjacency matrix, $\bm{A} \in \mathbb{R}^{|\mathcal{V}| \times |\mathcal{V}|}$, of a graph $\mathcal{G}$ represents the connections between the set of vertices.

        An induced subgraph of a graph $\mathcal{G}$ is the subgraph 
        formed by a subset of nodes $S \subseteq \mathcal{V}$ and all 
        edges whose endpoints are both in $S$. In this paper, a node 
        $v \in \mathcal{V}$ is selected uniformly at random at time 
        $t$; however, this done at this stage for proof of concept, ideally the DDDAS guides observation as decisions are made. A ball of radius $L$ centered at $v$, denoted 
        $\mathcal{B}(v, L)$, defines the $L$-hop neighborhood of $v$.
        
        This process repeats at each timestep $t \in \mathcal{T}$, where 
        the underlying graph $\mathcal{G}_t$ may itself vary across 
        timesteps, reflecting the dynamic nature of the instrumentation 
        system supplying the graph-structured data stream. Consequently, 
        both the node set $\mathcal{V}_t$ and edge set $\mathcal{E}_t$ 
        are time-dependent.

    \subsection{DDDAS and Decision-Making}

        The DDDAS concept integrates dynamic feedback between a computational model and its measurement and instrumentation system. This feedback is dependent on the task at hand. For this paper, the feedback is a decision-making process for classifying sequentially streamed graph-structured observations with a fixed set of classes $\mathcal{K}$.

        A measurement and instrumentation system streams graph-structured 
        data $\mathcal{G}_t$ at discrete time steps $t \in \mathcal{T}$. 
        At each timestep, a node $v_t \in \mathcal{V}_t$ is selected from the measured graph $\mathcal{G}_t$, according to an observation selection policy $\pi$, and the $L$-hop neighborhood $\mathcal{B}(v_t, L)$ is taken as an 
        observation $\bm{o}_t$. This observation selection is the instrumentation reconfiguration step in the DDDAS paradigm: $\pi$ determines which region of the physical system is interrogated at each subsequent timestep. In this work, $\pi$ is uniform over $\mathcal{V}_t$, so that the reported gains are attributable to the accumulation module rather than adaptive policy-making. 
        
        At each timestep, the computational model generates an evidence 
        vector $\bm{e}_t$ over the classes $\mathcal{K}$. The graph-structured data stream serves as the physical sensor and $\pi$ serves as the sensor management component of the DDDAS feedback loop. The evidence encoder and Bayesian filter serve as the estimation algorithm, and the posterior belief state is the modeled data maintained throughout the timesteps. B-EAGLe operates directly on measured graphs and does not instantiate a physical system simulation; the data augmentation loop is driven by the comparison of accumulated belief and the decision threshold. 

        Evidence accumulation is a decision-making paradigm. It handles uncertainty by deferring commitment until sufficient evidence is gathered \cite{agarwal2018better}. One such model takes inspiration from the Cortico-Basal Ganglia Thalamic Circuit (CBGT) loop that is responsible for decision-making and evidence accumulation \cite{wei2016inhibitory}. This model, the CBGT-Net \cite{sharma2024cbgtnetneuromimeticarchitecturerobust}, differs from traditional feed-forward networks as its input is a stream of observations. 
        
        In the original paper, the stream of observations is images. These images are run through an encoder at each time step. The encoder produces an evidence vector $\bm{e}_{t} = [e_{1,t}, e_{2,t}, \ldots, e_{K,t} ]$, where $K = |\mathcal{K}|$.
        
        Evidence is accumulated in an additive manner, with the accumulator 
        initialized as $\bm{a}_0 = \bm{0}$ and updated at each timestep as 
        $\bm{a}_t = \bm{e}_t + \bm{a}_{t-1}$. A decision is triggered once 
        any element of $\bm{a}_t$ exceeds a fixed integer threshold 
        $\tau \in \mathbb{Z}^{++}$. Thresholding is a feedback action on the instrumentation as crossing a threshold halts all observations for a given sensor. While this instantiates the data driven 
        update direction of the DDDAS feedback loop, the additive scheme 
        offers no probabilistic interpretation of the accumulated evidence. 
        B-EAGLe addresses this by extending CBGT-Net to graph-structured 
        instrumentation systems and replacing the additive accumulator with 
        a Bayesian filter, enabling principled uncertainty quantification 
        over the posterior belief state.



\section{Problem Formulation and Methodology}
    \label{sec:prob}
\begin{figure}
        \centering
        \includegraphics[width=0.9\textwidth]{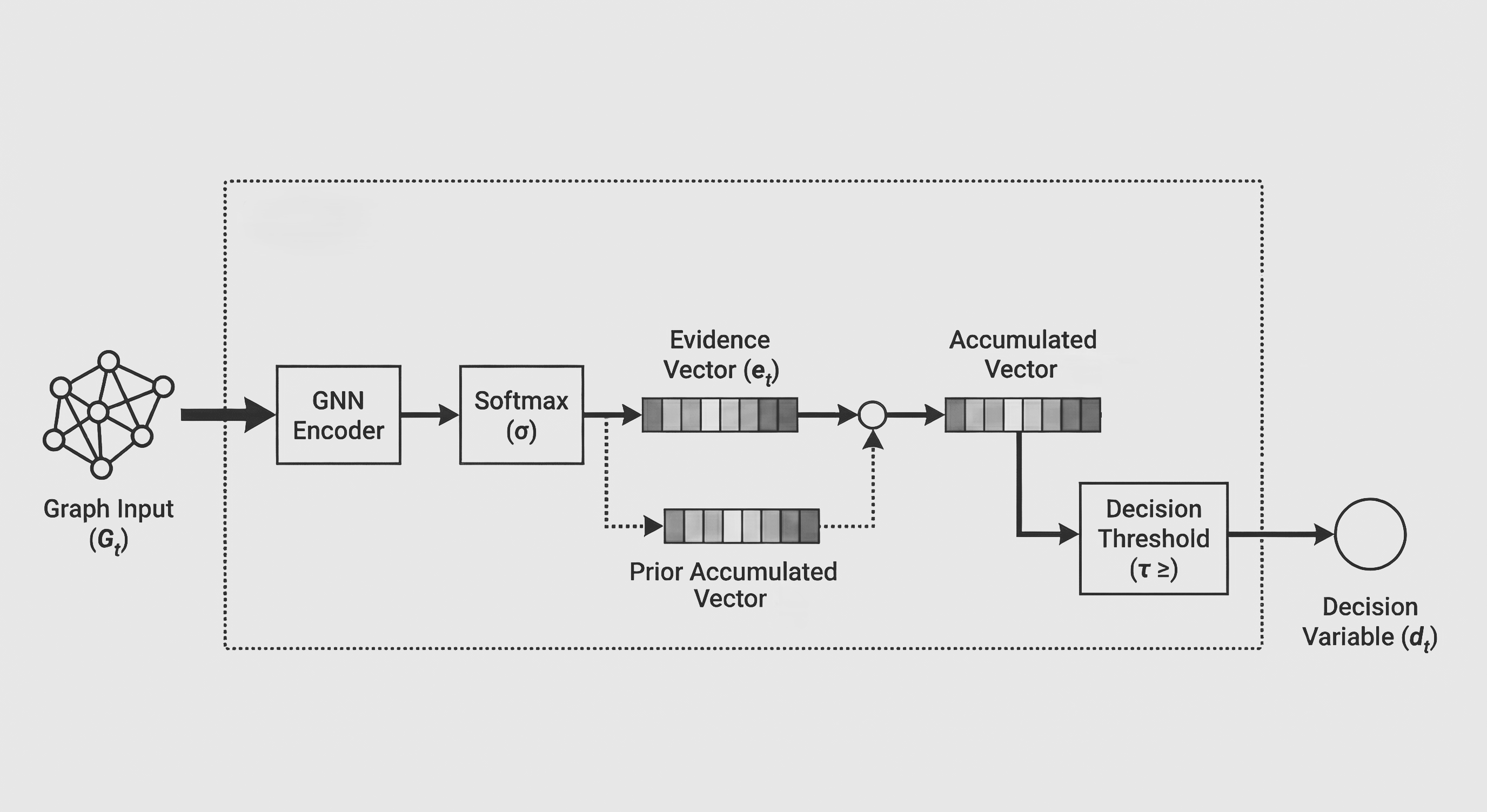}
        \caption{Overview of the B-EAGLe computational model illustrating the sequential evidence accumulation pipeline. The dashed border denotes the iterative loop that repeats until the decision variable $d_t = 1$ or $T_{\max}$ is reached.}
        \label{fig:B-Eagle}
    \end{figure}

    At each time step, a graph-structured data stream $\mathcal{G}_t$ is fed to the computational model by some type of instrumentation system. Due to computational complexity of running computations on whole graphs, an observation is an $L$-hop neighborhood induced subgraph of $\mathcal{G}$. This is the beginning step as shown in Fig.~\ref{fig:B-Eagle} with the rest of model depicted as well.

    The rest of the architecture is divided into three modules: the GNN encoder, the Bayesian Evidence Accumulation Module, and the Confidence Threshold Module.

    \subsection{GNN Encoder Module}
    
    At each time step $t$, the GNN encoder receives an observation $\bm{o}_t$ which is an $L$-hop subgraph of a random node in $\mathcal{G}_t$.
    The GNN encoder layers are based on the Graph Attention Network (GAT) \cite{veličković2018graphattentionnetworks}. The remaining architectural choices follow the GNN+ framework of \cite{luo2025}.
    
    The encoder incorporates edge feature integration into the 
    message-passing process, layer-wise batch normalization to 
    stabilize training, dropout to mitigate over-smoothing, residual 
    connections to prevent vanishing gradients, and a feed-forward 
    network appended at each layer to enable complex feature transformations beyond what the message-passing aggregation alone provides.
    
    For graph-level classification, global mean pooling aggregates node representations into a graph-level embedding. For node-level classification, this is replaced with a node-wise readout. The final embedding is passed through a softmax classifier producing $\bm{e}_t \in \Delta^{K-1}$, where 
    \[
      \Delta^{K-1} = \left\{ p \in \mathbb{R}^{K} \;\middle|\; \sum_{i=1}^{K} p_i = 1 \text{ and } p_i \geq 0 \;\forall i \right\}.
    \]

    \subsection{Bayesian Evidence Accumulation Module}
        The Bayesian Evidence Accumulation Module maintains the posterior distribution over the set of categories $\mathcal{K}$. This distribution is updated sequentially as new observations are made. Two assumptions underpin this approach. First, observations are conditionally independent given the true class. In other words, the current observation, $\bm{o}_t$, is independent of past observations, $\bm{o}_{1:t-1}$. Second, the prior is uniform at time $0$, $b_{k,0} = \frac{1}{K} \; \forall \; k \in  \mathcal{K}$. 

        The GNN Encoder Module, $E_\theta$, maps each $L$-hop neighborhood subgraph observation to a probability distribution over the classes via a softmax output. This probabilistic output for each class is
        \begin{equation}
            P(X=k \mid \bm{o}_t) = \frac{P(\bm{o}_t \mid X=k) \cdot P(X = k)}{P(\bm{o}_t)}.
        \end{equation}
        This is the posterior probability of class $k \, \in \mathcal{K}$ given the observation at time $t$. From Bayes' theorem, the observation likelihood can be expressed in terms of the posterior and the prior. From the second assumption, the prior is uniform at $t=0$: $b_{k,0} = \frac{1}{K} \; \forall \, k \in \mathcal{K}$. Under the conditional independence assumption, this uniform prior is applied at each timestep when computing the observation likelihood, such that $P(\bm{o}_t)$ does not depend on $k$.
        These two terms can be canceled, yielding the following:
        \begin{equation}
            P(\bm{o}_t \mid X = k) \propto e_{k,t},
        \end{equation}
        where $e_{k,t}$ is the softmax output of the GNN encoder for class $k$ at time $t$. This permits the sequence of encoder outputs to be treated as normalized observation likelihoods, enabling the accumulation module to perform Bayesian filter inference.

        The module updates using a Bayesian filter,
        \begin{equation}
            b_{k,t} = \frac{(e_{k,t} + \varepsilon) \cdot (b_{k,t-1} + \varepsilon)}
            {\sum_{j=1}^{K} (e_{j,t} + \varepsilon) \cdot (b_{j,t-1} + \varepsilon)},
            \label{eq:bayesianfilter}
        \end{equation}
        where $b_{k,t}$ is the posterior belief that the input belongs to class $k$ given all observations up to time $t$, $e_{k,t}$ is the likelihood estimate produced by the GNN encoder at time $t$, and $\varepsilon > 0$ is a small constant that prevents zero-probability terms, consistent with 
        Cromwell's Rule. 
        The multiplicative update suppresses classes assigned low 
        likelihood across successive observations, while the normalization denominator ensures $\sum_{k=1}^{K} b_{k,t} = 1$ at every timestep.

         The update in Eq.~\eqref{eq:bayesianfilter} operates under the assumption of the observations being conditionally independent. However, in instances where observations are made in overlapping regions, particularly in single graph datasets, the observations would be correlated. The effect is that a partially overlapping observation contributes evidence already reflected in the belief state, thus the posterior concentrates more rapidly than what is warranted from the information given by the observation sequence. The accumulated belief state is therefore overconfident in a classification and the stopping criteria may be satisfied sooner than what is justified by the evidence. A sufficiently stringent $\tau$ offsets this effect by requiring more accumulated evidence before a decision is committed. However, an overly stringent of a $\tau$ target may contribute to high timeout rates reported in Section~\ref{tab:results}. 

    \subsection{Confidence Threshold Module}

        The original CBGT-Net architecture deploys an additive accumulation module with a fixed decision threshold, $\tau \in \mathbb{Z}^{++}$, triggering a decision once one of the classes exceeds it. Instead, B-EAGLe operates on the posterior belief state, $b_{k,t}$. This enables a probabilistic stopping criterion rooted in a probabilistic confidence threshold. A decision is triggered at the first time step in which
        \begin{equation}
            d_t = \vec{1} \big[ \max_k \, b_{k,t} \geq \tau \, \big] \qquad \tau \in (0,1),
        \end{equation}
        where $d_t$ is the binary stopping decision once one of the likelihoods meets or exceeds $\tau$ and where $\tau$ is the confidence threshold hyperparameter given to the model.
        The predicted class is given by
        \begin{equation}
            \hat{y} = \argmax_{k \in \mathcal{K}} \, b_{k,t_d},
        \end{equation}
        where $t_d = \min\{t : d_t = 1\}$ is the first timestep at which 
        the confidence threshold is met.

        The stopping criterion partially fulfills the DDDAS concept by the model continually collecting evidence while $d_t = 0$, and when $d_t =1$, data acquisition halts and the classification is committed.


\section{Case Study and Results}
    \label{sec:case}
        This paper uses widely adopted and easily available Super-pixel datasets for graph classification tasks, MNIST-SP and CIFAR10-SP, and Stochastic Block Model datasets for node classification tasks, PATTERN and CLUSTER, as presented in \cite{dwivedi2022benchmarkinggraphneuralnetworks}.

        For the graph classification datasets, 3-hop neighborhoods are used to capture broader structural context across the Super-pixel graph. For the node classification datasets, 1-hop neighborhoods are used as the local neighborhood structure is sufficient for node-level classification.

        Hyperparameter configurations follow the GNN+ framework of 
        \cite{luo2025}, including edge feature integration, batch 
        normalization, dropout, residual connections, and feed-forward 
        networks. The confidence threshold was set to $\tau = 0.99$ for all experiments, meaning a decision is triggered when the posterior belief of any class reaches or exceeds 99\%.
        Following the implementations of \cite{sharma2024cbgtnetneuromimeticarchitecturerobust}, the number of time steps permitted for evidence collection is $T_{\max} = 50$. Each experiment is run through multiple seeds and the standard deviation is given for the reported results. Episodes that do not reach the confidence threshold within $T_{\max}$ are classified by the $\argmax$ of the final timestep. Additional hyperparameters can be found in Table~\ref{tab:hyperparams}.

        \begin{table}
            \caption{Hyperparameter settings of B-EAGLe following \cite{luo2025}.}
            \label{tab:hyperparams}
            \centering
            \begin{tabular}{|l|c|c|c|c|}
            \hline
            \textbf{Hyperparameter} & \textbf{MNIST-SP} & \textbf{CIFAR10-SP} & \textbf{PATTERN} & \textbf{CLUSTER} \\
            \hline
            \# GAT Layers        & 6      & 5      & 12       & 12       \\
            \# $L$-hops            & 3      & 3      & 1        & 1        \\
            Edge Features  & True   & True   & True     & False    \\
            Dropout              & 0.15   & 0.05   & 0.05     & 0.1      \\
            Hidden Dim           & 60     & 65     & 90       & 90       \\
            Graph Pooling        & mean   & mean   & readout  & readout      \\
            \hline
            Batch Size           & 16     & 16     & 32       & 16       \\
            Learning Rate        & 0.0005 & 0.001  & 0.001    & 0.001    \\
            \# Epochs            & 200    & 200    & 200      & 100      \\
            Weight Decay         & 1e-5   & 1e-5   & 1e-5     & 1e-5     \\
            \hline
            Confidence $\tau$    & 0.99   & 0.99   & 0.99     & 0.99     \\
            $T_{\max}$           & 50     & 50     & 50       & 50       \\
            \hline
            \end{tabular}
        \end{table}

        As shown in Table~\ref{tab:results}, the B-EAGLe performed better than the state-of-the-art baselines for three out of the four datasets. Although MNIST-SP is an easy dataset, the results suggest that B-EAGLe's sequential evidence accumulation provides greater benefit on datasets requiring more complex decision boundaries.
        
        \begin{table}[pbt!]
        \caption{Classification accuracy (\%) of B-EAGLe compared against state-of-the-art baselines from \cite{luo2025}. The average number of observations acquired before committing to a decision is included. The Timeout rate denotes the fraction of episodes requiring all $T_{\max}$ observations. Highlighted in green are the best results per dataset.}\label{tab:results}
        \centering
        \begin{tabular}{|l|c|c|c|c|c|}
        \hline
        \textbf{Dataset} & \textbf{Best Baseline} & \textbf{Baseline Acc.} & 
        \textbf{B-EAGLe} & \textbf{Avg.~$t_d$} & \textbf{Timeout} \\
        \hline
        MNIST-SP   & GatedGCN+ & \colorbox{green}{$98.71 \pm 0.137$} & $98.22 \pm 0.18$ & $11.0 \pm 0.16$ & 0.4\% \\
        CIFAR10-SP & GatedGCN+ & $77.22 \pm 0.381$ & \colorbox{green}{$90.13 \pm 2.75$} & $26.6 \pm 0.33$ & 1.6\%  \\
        PATTERN & GRIT & $87.20 \pm 0.076$ & \colorbox{green}{$94.00 \pm 4.39$} & $50.98 \pm 0.01$ & 100\%  \\
        CLUSTER & GRIT & $80.03 \pm 0.277$ & \colorbox{green}{$82.34 \pm 0.11$} &  $50.85 \pm 0.05$  & 98.6\%  \\
        \hline
        \end{tabular}
        \end{table}

        For graph classification datasets like MNIST-SP and CIFAR10-SP, the high confidence threshold did not lead to high timeout rates.
        However, for the node classification datasets, the timeout percentage neared 100\%. This suggests that $\tau=0.99$ can be too stringent.

\section{Conclusion and Future Work}
    \label{sec:conclusion}
    B-EAGLe has demonstrated that it performs better than the state-of-the-art baselines. On CIFAR10-SP the improvement is approximately 13 percentage points, obtained under an observation budget capped at 50 views per graph; PATTERN improves by almost 7 percentage points and CLUSTER by 2. The node-level benchmarks rarely satisfied the high confidence threshold at $\tau = 0.99$. The seed-to-seed variance on CIFAR10-SP is higher than the baseline, and this potentially attributed to the variance of the predicted class rather than the accumulation module. 

    B-EAGLe is also geared towards DDDAS implementation. In a complete DDDAS system, the computational model would not only improve via observations streamed, but would also steer which observations are acquired next. Although this paper exceeds the aforementioned baselines on three out of the four tested datasets, it does so with only one direction of the feedback loop fully realized.

    For future work, connecting B-EAGLe to a live measurement and instrumentation system would evaluate decision latency in a real-time dynamic setting. In such environments, committing to a decision quickly can be as valuable as an accurate prediction. Improving the data selection process during low-certainty episodes would enable the instrumentation system to prioritize informative observations and reduce the contribution of low-fidelity data on the belief state. Additionally, the high timeout rates on PATTERN and CLUSTER suggest that a sensitivity analysis over $\tau$, or potential need for a dynamic threshold module. Either would clarify whether a less stringent criterion recovers adaptive behavior on node-level tasks. A calibration analysis of the accumulated belief state would further establish whether the posterior supports the confidence interpretation placed on it. Finally, comparisons against baselines ensembled over a matched observation budget would separate the contribution of the recursive Bayesian update from that of multi-view aggregation alone.

\begin{credits}
\subsubsection{\ackname}
This work was supported in part by the U.S. Army DEVCOM Army Research Laboratory (ARL) under contracts W911NF-22-2-0143, W911NF-21-20186.

\subsubsection{\discintname}
The authors have no competing interests to declare that are relevant to the content of this article.


\end{credits}
%
%
%
\bibliographystyle{splncs04}
\bibliography{mybibliography}

@inproceedings{fan2019graphsocial,
  title={Graph neural networks for social recommendation},
  author={Fan, Wenqi and Ma, Yao and Li, Qing and He, Yuan and Zhao, Eric and Tang, Jiliang and Yin, Dawei},
  booktitle={The world wide web conference},
  pages={417--426},
  year={2019}
}

@article{li2023surveysocial,
  title={A survey of graph neural network based recommendation in social networks},
  author={Li, Xiao and Sun, Li and Ling, Mengjie and Peng, Yan},
  journal={Neurocomputing},
  volume={549},
  pages={126441},
  year={2023},
  publisher={Elsevier}
}

@article{pavlopoulos2011using,
  title={Using graph theory to analyze biological networks},
  author={Pavlopoulos, Georgios A and Secrier, Maria and Moschopoulos, Charalampos N and Soldatos, Theodoros G and Kossida, Sophia and Aerts, Jan and Schneider, Reinhard and Bagos, Pantelis G},
  journal={BioData mining},
  volume={4},
  number={1},
  pages={10},
  year={2011},
  publisher={Springer}
}

@article{lesne2006complexbio,
  title={Complex networks: from graph theory to biology},
  author={Lesne, Annick},
  journal={Letters in Mathematical Physics},
  volume={78},
  number={3},
  pages={235--262},
  year={2006},
  publisher={Springer}
}

@misc{sharma2024cbgtnetneuromimeticarchitecturerobust,
      title={CBGT-Net: A Neuromimetic Architecture for Robust Classification of Streaming Data}, 
      author={Shreya Sharma and Dana Hughes and Katia Sycara},
      year={2024},
      eprint={2403.15974},
      archivePrefix={arXiv},
      primaryClass={cs.NE},
      url={https://arxiv.org/abs/2403.15974}, 
}

@misc{luo2025,
      title={Can Classic GNNs Be Strong Baselines for Graph-level Tasks? Simple Architectures Meet Excellence}, 
      author={Yuankai Luo and Lei Shi and Xiao-Ming Wu},
      year={2025},
      eprint={2502.09263},
      archivePrefix={arXiv},
      primaryClass={cs.LG},
      url={https://arxiv.org/abs/2502.09263}, 
}

@article{sahili2023spatio,
  title={Spatio-temporal graph neural networks: A survey},
  author={Sahili, Zahraa Al and Awad, Mariette},
  journal={arXiv preprint arXiv:2301.10569},
  year={2023}
}

@article{zheng2025survey,
  title={A survey of dynamic graph neural networks},
  author={Zheng, Yanping and Yi, Lu and Wei, Zhewei},
  journal={Frontiers of Computer Science},
  volume={19},
  number={6},
  pages={196323},
  year={2025},
  publisher={Springer}
}

@article{khoshraftar2024survey,
  title={A survey on graph representation learning methods},
  author={Khoshraftar, Shima and An, Aijun},
  journal={ACM Transactions on Intelligent Systems and Technology},
  volume={15},
  number={1},
  pages={1--55},
  year={2024},
  publisher={ACM New York, NY}
}

@article{xia2021graph,
  title={Graph learning: A survey},
  author={Xia, Feng and Sun, Ke and Yu, Shuo and Aziz, Abdul and Wan, Liangtian and Pan, Shirui and Liu, Huan},
  journal={IEEE Transactions on Artificial Intelligence},
  volume={2},
  number={2},
  pages={109--127},
  year={2021},
  publisher={IEEE}
}

@article{ouyang2007dynamic,
  title={Dynamic data driven application system: Recent development and future perspective},
  author={Ouyang, Ying and Zhang, Jia En and Luo, Shi Ming},
  journal={Ecological modelling},
  volume={204},
  number={1-2},
  pages={1--8},
  year={2007},
  publisher={Elsevier}
}

@misc{dwivedi2022benchmarkinggraphneuralnetworks,
      title={Benchmarking Graph Neural Networks}, 
      author={Vijay Prakash Dwivedi and Chaitanya K. Joshi and Anh Tuan Luu and Thomas Laurent and Yoshua Bengio and Xavier Bresson},
      year={2022},
      eprint={2003.00982},
      archivePrefix={arXiv},
      primaryClass={cs.LG},
      url={https://arxiv.org/abs/2003.00982}, 
}

@inproceedings{darema2005dddas,
  title={Dynamic data driven applications systems: New capabilities for application simulations and measurements},
  author={Darema, Frederica},
  booktitle={International conference on computational science},
  pages={610--615},
  year={2005},
  organization={Springer}
}

@article{Jiang_2022,
   title={Graph neural network for traffic forecasting: A survey},
   volume={207},
   ISSN={0957-4174},
   url={http://dx.doi.org/10.1016/j.eswa.2022.117921},
   DOI={10.1016/j.eswa.2022.117921},
   journal={Expert Systems with Applications},
   publisher={Elsevier BV},
   author={Jiang, Weiwei and Luo, Jiayun},
   year={2022},
   month=Nov, pages={117921} }

@misc{zhang2018graphedgeconvolutionalneural,
      title={Graph Edge Convolutional Neural Networks for Skeleton Based Action Recognition}, 
      author={Xikun Zhang and Chang Xu and Xinmei Tian and Dacheng Tao},
      year={2018},
      eprint={1805.06184},
      archivePrefix={arXiv},
      primaryClass={cs.CV},
      url={https://arxiv.org/abs/1805.06184}, 
}

@misc{veličković2018graphattentionnetworks,
      title={Graph Attention Networks}, 
      author={Petar Veličković and Guillem Cucurull and Arantxa Casanova and Adriana Romero and Pietro Liò and Yoshua Bengio},
      year={2018},
      eprint={1710.10903},
      archivePrefix={arXiv},
      primaryClass={stat.ML},
      url={https://arxiv.org/abs/1710.10903}, 
}

@article{schizas2015dynamic,
  title={Dynamic data driven sensor network selection and tracking},
  author={Schizas, Ioannis D and Maroulas, Vasileios},
  journal={Procedia Computer Science},
  volume={51},
  pages={2583--2592},
  year={2015},
  publisher={Elsevier}
}

@article{cencerrado2011prediction,
  title={Prediction Time Assessment in a DDDAS for Natural Hazard Management: Forest Fire Study CaseI},
  author={Cencerrado, And{\'r}es and Cort{\'e}s, Ana and Margalef, Tom‘as},
  journal={Procedia Computer Science},
  volume={4},
  pages={1761--1770},
  year={2011},
  publisher={Elsevier}
}

@inproceedings{madey2007enhanced,
  title={Enhanced situational awareness: Application of DDDAS concepts to emergency and disaster management},
  author={Madey, Gregory R and Barab{\'a}si, Albert-L{\'a}szl{\'o} and Chawla, Nitesh V and Gonzalez, Marta and Hachen, David and Lantz, Brett and Pawling, Alec and Schoenharl, Timothy and Szab{\'o}, G{\'a}bor and Wang, Pu and others},
  booktitle={International conference on computational science},
  pages={1090--1097},
  year={2007},
  organization={Springer}
}

@article{agarwal2018better,
  title={Better safe than sorry: Evidence accumulation allows for safe reinforcement learning},
  author={Agarwal, Akshat and Kumar V, Abhinau and Dunovan, Kyle and Peterson, Erik and Verstynen, Timothy and Sycara, Katia},
  journal={arXiv preprint arXiv:1809.09147},
  year={2018}
}

@article{wei2016inhibitory,
  title={Inhibitory control in the cortico-basal ganglia-thalamocortical loop: complex regulation and interplay with memory and decision processes},
  author={Wei, Wei and Wang, Xiao-Jing},
  journal={Neuron},
  volume={92},
  number={5},
  pages={1093--1105},
  year={2016},
  publisher={Elsevier}
}

\end{document}